*Neural Networks*

# Dynamic Action Recognition: A convolutional neural network model for temporally organized joint location data

Adhavan Jayabalan[1], Harish Karunakaran[1], Shravan Murlidharan[1], Tesia Shizume[2]

[1]Department of Robotics Engineering, Worcester Polytechnic Institute, Worcester, MA, USA

[2]Department of Computer Science, Worcester Polytechnic Institute, Worcester, MA, USA.

**Abstract**
**Motivation:** Recognizing human actions in a video is a challenging task which has applications in various fields. Previous works in this area have either used images from a 2D or 3D camera. Few have used the idea that human actions can be easily identified by the movement of the joints in the 3D space and instead used a Recurrent Neural Network (RNN) for modeling. Convolutional neural networks (CNN) have the ability to recognise even the complex patterns in data which makes it suitable for detecting human actions. Thus, we modeled a CNN which can predict the human activity using the joint data. Furthermore, using the joint data representation has the benefit of lower dimensionality than image or video representations. This makes our model simpler and faster than the RNN models. In this study, we have developed a six layer convolutional network, which reduces each input feature vector of the form 15x1961x4 to an one dimensional binary vector which gives us the predicted activity.
**Results:** Our model is able to recognise an activity correctly upto 87% accuracy. Joint data is taken from the Cornell Activity Datasets which have day to day activities like talking, relaxing, eating, cooking etc.
**Availability:** A python implementation of our model is available at https://github.com/jadhavan/AI-Project.
**Contact:** ajayabalan@wpi.edu

## 1 Introduction

As an important branch of computer vision, action recognition has applications in video surveillance, robot vision, human-computer interaction etc. With the ever growing presence of robots and other automated services in people's lives, computer recognition of human actions is of ever growing importance. It is important to enable these robots and services to be more aware of the human factor in the environment. This can help prevent accidents in settings where there is close human- robot interaction like factories or in the future: on roadways, in senior or other extended care facilities, in the home, or in education. Furthermore, as more people take increasing advantage of such services and have increased interactions with robots, there is an increasing demand for more performant techniques.

Recognizing human actions from a live video feed is a challenging task as there are many factors that can change the task environment making it partially observable. These factors include background noise, viewpoint, lighting and so on. Moreover, the subjects also differ in the way they perform an action. At present, the algorithm developed by Google Research can recognize simple human actions and complex ones up to an accuracy of around 60%. They have addressed this problem by using the temporal connectivity pattern in a CNN architecture and its effects on performance on the addition of new motions, in an empirical way [1]. This paper provides a method to improve the classification of complex human actions from a live video feed by using the joint positions of a human in a Convolutional Neural Network (CNN).

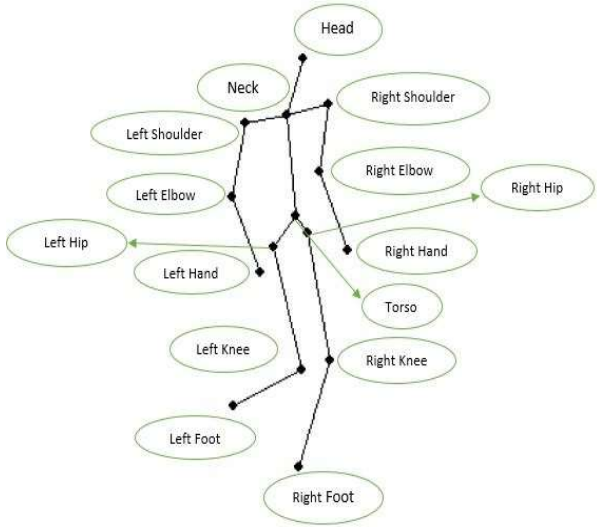

**Figure 1: Schematic Representation of Human Joints.** Each node has x, y, z spatial coordinates.

Previous work in the area of action recognition and detection from images and joints have only been recorded by 2D cameras. But, in the general case, human actions can only be represented in the 3D space. The human body can be treated as an articulated system with rigid bones and hinged joints which are connected to the four limbs and the trunk [2]. The human body can be divided into the following fifteen joints, Head, Neck, Torso, Left Shoulder, Left Elbow, Right Shoulder, Right Elbow, Left Hip, Left Knee, Right Hip, Right Knee, Left Hand, Right Hand, Left Foot and Right Foot as shown in Figure 1. Each joint is represented by its position in 3D space. The joint positions are also given a confidence value of either zero or one to indicate the accuracy of the joint prediction [3]. These form the features of every frame of the video.

Convolutional Neural Networks are very successful at static image recognition problems such as the MNIST dataset. CNN's are able to automatically learn complex features for object recognition instead of using hand-crafted features. This makes the CNN an ideal candidate for video and action classification tasks. Video analysis is more informative for action classification because it includes a temporal component which allows the use of motion and other features which can be used in the neural network. But at the same time, as the number of features increases and the number of frames in a video is typically in the thousands, the computational complexity can increase drastically. Thus, the naïve approach of using each video frame as an image and applying CNN would contain incomplete information and the complexity can exceed present computational capacity for large video samples [4] (eg. Youtube videos).

Also, the modelling of the temporal evolution of the video which is needed for its accurate classification is difficult because of the variable length of videos. The approach which can be used to solve this problem is feature pooling, where each frame is processed with a CNN and then all the frame information or features are combined using pooling layers to give the final output. By the CNN architecture, since there is sharing of parameters between each layer and thereby between frames, we are able to maintain the constant number of parameters when the description of the video's temporal evolution is obtained. In our case, each frame would contain the position information of the fifteen joints defined by the data set [4].

Finding the quantity of the data for the articulated joint position and orientation which would be needed to achieve significant testing results is a major challenge. The use of joints to represent the human body does not represent the complete information about the action performed by a human. Therefore, some preprocessing is required to train the model to detect joints in the presence of uncertainties. There have been various approaches to solve the problem of action recognition using joint parameters but most of the work have used Recurrent Neural Network (RNN) with Long-Short Term Memory (LSTM) or Hidden Markov Models (HMM). We propose to use a CNN for action recognition from joint parameters.

The general problem addressed in this paper is the classification of videos of humans performing complex actions into their actions. The approach is to reduce the dimensionality of the problem by representing each frame in each video as a set of human joint positions. The data used for the paper are the Cornell Activity Datasets (CAD) which contain joint positions and orientations of fifteen joints in the human body. Sample images from the datasets are shown in Figure 2. This data became the feature vector for each video and for classifying the videos, a time-domain convolutional neural network is trained on the data and tested which the ratio of the train to test data as 4:1.

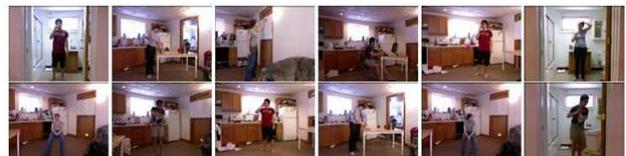

**Figure 2: Samples from the CAD Datasets.** Row-wise, from left: brushing teeth, cooking (stirring), writing on whiteboard, working on computer, talking on phone, wearing contact lenses, relaxing on a chair, opening a pill container, drinking water, cooking (chopping), talking on a chair and rinsing mouth with water [9].

The rest of the paper is divided as follows, the review of the related work is in Section 2. The overview of our methodology and neural network is in Section 3. Then, the results are discussed in Section 4 and the finally the conclusion is in Section 5.

## 2 Related Work

Recognition of human actions in videos is a challenging task which has received a significant amount of attention in the research community. Compared to still image classification, the temporal component of videos provides an additional feature for recognition, as a number of actions can be recognized more accurately based on motion information of a human. In addition, video provides natural data augmentation for single image classification. Video recognition research has been largely driven by the advances in image recognition methods, which were often adapted and extended to deal with video data. But such methods involve intensive computational complexity which makes them not suitable for large datasets.

Several methods have addressed the issue of estimation of human pose using a neural network which is an approach which removes the problem of occluded joints when a computer vision approach is employed. Toshev et al. [5] have estimated the human pose using a Deep Neural Network. The authors use a novel method using DNN's for visual classification and precise localization of articulated objects to the problem of human pose. Their method is able to capture the full context of each body joint and that a cascade of DNN- based pose predictors gives increased precision of joint locations. They are also able to achieve a running time of 0.1 seconds per image on a 12-core CPU. Despite the fast computation speed, their DNN-based regression method is still unable to give precise joint localization when the joints are occluded which was one of the problems of the earlier computer vision approach.

Baccouche et al. [6] developed a fully automated two-step neural-based deep model for human action recognition without any prior knowledge. The first part extends the CNN to 3D case thereby learning spatio-temporal features. Then, using the learned features a recurrent neural network is used to classify the video. This automated steps removes the need for hand-crafted features for every specific task which can be highly problem dependent. They achieved an overall accuracy of 94.39% on the KTH dataset on cross validation with five runs. However, their method, since it involves a 3D Convolution has very high computation time which is a major drawback. Feichtenhofer et al. [7] use a novel spatiotemporal architecture with a convolutional and temporal fusion layer to incorporate motion information for both appearance in still images and stacks of optical flow to mitigate the problem of occlusion of joints which occur frequently in real-life scenarios. They extend the work on the two-stream convolutional network for action recognition in videos by Simoyan and Zisserman [14] which had the problem of pixel-wise correspondences between spatial and temporal features. When the method is tested on the UCF-101 data, they report a 4.5% improvement in recognition when using a VGG-16 for both streams.

However, these results can further be improved by combining Convolutional Network predictions with FV-encoded IDT features. Since all these approaches uses the spatio-temporal architecture, the computational time can become a big problem. The concept introduced by Bilen et al. [8] on the compact representation of videos which can be used in CNN's called dynamic images (RGB image which summarizes the whole video in a compressed format) would be good solution to the above problem. To achieve this, they build the dynamics of a video directly from the image pixels instead of an intermediate feature representation. This reduces the complexity, compression factor and the efficiency of the CNN whereas capturing the dynamic image. They report a 2-3% accuracy improvements in the UCF-101 dataset and that dynamic images pooled on top of static RGB frames lead to better results. In order to improve the accuracy, the authors propose combining the dynamic images with sophisticated encodings.

Hei Ng et al. [4] propose two CNN architectures to process individual video frames, AlexNet and GoogLeNet. They investigate LSTM networks capable of learning from temporally ordered sequences and various feature pooling architectures. GoogLeNet stacks Inception modules to form a network of 22 layers. It outperforms Karpathy et al. [1] by a margin of 4.3-5.6%.

Several methods exist to recognize actions by various methods. Sung et al. [9] have devised a method to sub-sequences of the various activities using a two-layered maximum-entropy Markov model. They use a supervised learning method approach where the ground truth of the labeled data is given for training. They have reported an 84.3% accuracy for activity recognition on their dataset. The strength of their method is that it reports a neutral value most of the time which reduces the prediction error and also learn new actions. But, their method cannot be used when there is any occlusion and it cannot recognize actions where the human interacts with the environment. Meanwhile, Gupta et al. [10] in his work has shown that modelling of interaction between human poses and objects in a 2D video gives better performance for the tasks of object detection and activity recognition. But, it does not take into account the 3D relations between activity and the data and also the reduction in quality when using 2D data.

Several approaches to the action recognition problem have used the human skeletal structure to predict actions. Although, these works have significant accuracy, their approaches are based on Recurrent Neural Network (RNN) and Hidden Markov Models (HMM) which don't use the inherent advantages of using a CNN which removes the problem of overfitting and has shown to significantly increase the accuracy of other image classification methods. Yong Du et al. [11] divide the human skeleton into five parts according to the human physical structure and use five separate neural networks. They use a novel hierarchical bidirectional RNN to recognize the action of a human and compare it with four other architectures on the MSR Action 3D dataset. Their hierarchical RNN architecture outperforms all the other four architectures. But, one problem in their method arises from the overfitting and underfitting during training. Also, it becomes difficult to distinguish human actions only from the skeletal joints. Thus, the presence of more features in the recurrent neural network would improve the precision of human action recognition.

However, the use of RNN can lead to overfitting when the number of features is less when dealing with temporal data. Woo Young Kwon et al. [12] developed an improved skeleton tracker and complexity based motion features by combining the joint positions from a Kinect sensor and a Kalman filter framework. They used a supervised learning with a deep recurrent neural network for this purpose. They use their own dataset which consists of 16 activity classes to test their method. They are able to achieve an accuracy of 70% which is around 11% greater than other baseline methods. But, their method is computationally complex since it involves computing the Kalman Filter at each time step of the video. Di Wu and Ling Shao [13] use a hierarchical dynamic framework to extract the high level skeletal joint features and use deep neural networks to predict the probability distribution over the states of the Hidden Markov model. They choose a HMM due to the fact that an RNN tends to overemphasize the temporal information in the presence of insufficient data which can lead to overfitting. Their work is able to achieve better results on the MSRC 12 dataset when compared to Randomized Forest and Structured Streaming Skeletons with an F-score increase of around 0.01, but still cannot address the issue of overfitting which is present in RNN.

## 3 Methods

One of the general machine learning problems is that, videos of performing complex actions cannot be accurately classified into their actions.This work attempts to do so. The data chosen for this work was the Cornell Activity Dataset (CAD) [20, 23, 9]. The approach taken in this work was to reduce the dimensionality of the problem by representing each frame in each video as a set of human joint positions. This data became the feature vector for each video. Then to classify the videos, a neural network was trained on the data.

### 3.1 Data Selection and Preprocessing

Given the focus of the project, the qualities we looked for in the selection of a dataset are: action labels, actions of appropriate complexity or subtlety, a sufficient variety of actions, and the size of the dataset. The presence of labels and the size of the dataset were important because we were using a supervised learning strategy [17]. That the dataset had actions of appropriate variety and complexity or subtlety were important to ensure that the data was aligned with the task. Using these criteria, we considered several means of obtaining data for our project as well as existing datasets. Direct collection was ruled out due to the time constraints and it was out of the scope of the project. We also investigated using a joint detector on an existing video dataset, however we did not find one that performed to our standards. This left us with finding an existing dataset that had the qualities we were looking for as well as joint information. The final candidates for a dataset where the Cornell Activity Dataset (CAD) [20, 23, 9] and the Frames Labeled In Cinema (FLIC) dataset [25]. We chose CAD over FLIC because the actions in CAD were more complicated than the actions in FLIC.

The Cornell Activity Dataset is comprised of two sections. CAD-60 is the older portion and is comprised of data obtained using a Microsoft Kinect to acquire RGBD data from which they extracted skeletal data using a tracking system provided by PrimeSense [9]. The orientation subcomponent of which they then transformed to be relative to the person's torso, instead of the sensor. CAD-60 is a well-studied dataset using various data representations. However, many of the methods that have been applied to the dataset use variations on Markov Models [22, 9] or use SVMs for the classification portion of the problem [4, 18, 21, 22, 27], leaving opportunities to investigate the effectiveness of other classification algorithms and techniques. CAD-120 is also comprised of data obtained using a Microsoft Kinect but uses Openni's skeleton tracker, which they mention provides somewhat noisy data [23].

We decided to use both sections of the CAD for a larger dataset. As a consequence, data consolidation and normalization decisions needed to be made. Some of the action categories were very similar, thus to reduce the chance of complicating the learning these similar action categories were joined. The action categories "talking on the phone" and "talking on phone" were combined into the more general category of "talking". Similarly, "eating" and "cereal" were combined into "eating" and "stacking", "unstacking" and "placing" were combined into "stacking". The rest of the action categories remained the same.

Furthermore, to the number of frames in each video was not consistent across videos. Thus, a normalization method was required to ensure the feature vectors would be the same size. Several options were considered (Table 1). The Repetition technique of normalizing the number of frames in the feature vector was used for its simplicity and distribution of frame repetition across multiple frames.

**Table 1: Comparison of Data Normalization Techniques.** Projected benefits and drawbacks are compared among four different methods for normalizing the number of frames in the feature vector

| Method | Projected Benefits | Projected Drawbacks |
|---|---|---|
| Truncation | ● Avoids transposing frames to temporally disparate locations<br>● Simple to Perform | ● Reduced data to process and learn<br>● Terminal information may be important<br>● Temporal duration lost? |
| Padding (Terminal Data) | ● Avoids transposing frames to temporally disparate locations<br>● Indirect measure of temporal duration<br>● Simple to Perform | ● May overemphasize the last frame (or first frame)<br>● Temporal duration lost? |
| Repetition | ● Minimizes the number of times any one frame is repeated<br>● Simple to Perform | ● Transposes frames to temporally disparate locations<br>● Temporal duration lost? |
| Padding (Special Value) | ● Avoids transposing frames temporally disparate locations<br>● Indirect measure of temporal duration | ● Avoids transposing frames temporally disparate locations<br>● Temporal duration information may exacerbate overfitting<br>● Determination and Validation of the Special Value becomes a confounding factor |

### 3.1.1 Adding Gaussian Noise

Even after combining both the CAD sections and using other pre-processing techniques, the population of some of the classes in the dataset were not large enough for the neural network to train. For populating those classes we added Gaussian noise to the dataset with zero mean and different standard deviations varying from 0.1 to 0.5 and created five different datasets to study its effects. The features of the new data was obtained using Equation 1 below,

$$f_{i,new} = f_{i\ old}^{j} + N(0, \sigma) \quad (1)$$

where,
$f_{i,new}$ = i[th] feature of a new sample
$f_{i\ old}^{j}$ = i[th] feature of the j[th] sample
$N(0, \sigma)$ = Gaussian noise with zero mean and standard deviation $\sigma \epsilon [0.1, 0.5]$

### 3.2 Neural Network

A Neural Network is a supervised machine learning approach that has been shown to be effective in Computer Vision classification tasks, and is potentially more resistant to lower magnitudes of data than Support Vector Machines [16]. This is especially true for Convolutional Neural Networks (CNNs) [4]. Neural Networks work by creating a model composed of many subcomponent

"neuron" layers. Each neuron takes an input and processes it via an activation function to produce an output, there may be additional operations that the neuron does on its input but this is the generic case. A weighted function of the outputs of one layer of neurons can then be passed to each neuron in the next layer of the neural network. To be of use, the completed model must be trained on a dataset. In training, the model uses data organized as feature vectors find the weights relevant to each layer by using an optimizer. An optimizer seeks to either minimize the difference or maximize the similarities between the predictions the model creates and the actual values corresponding to the class labels of each data in the training set.

The different types of layers in a neural network take names depending on how they process or how connected they are with the previous layer. Similarly the names of types of neural networks reflect the selection of layer types found in an instance of the type. CNNs utilize one or more convolutional layers to process information and find features in data. A convolutional layer applies a convolution to the input of the layer prior to assessing the activation function. Convolution is a mathematical technique common in signal processing. It can be thought of as cumulative effect of passing a local filter across the signal. In the case of common CNNs the signal is the image and is in 2D. Whereas CNNs for still images frequently convolve across the image plane, this work uses convolution to find features across the joint-position, time plane.

The design of this work's neural network was inspired and informed by "Beyond Short Snippets: Deep Networks for Video Classification" [4]. In that paper, Yue-Hei Ng et al. compared various methods of dealing with the time component of the problem of video action categorization. Where in the paper the Yue-Hei Ng et al. used stacked convolutional layers, in this work joint position data is used. Furthermore, this work focuses on a method most similar to the Time-Domain Convolution mentioned in that paper, as shown in Figure 3.

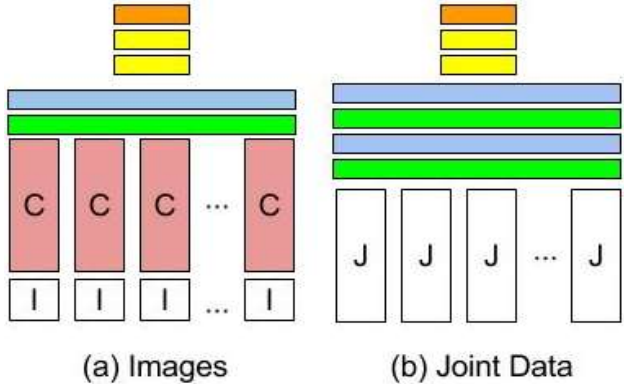

**Figure 3: Different Input Data Representations.** Input data representation is in white. 'J' indicates data represented as joint coordinates. 'I' indicates data represented as images. Red, Blue, Green, Yellow, and Orange represent image-plane convolutional layer stacks, max-pooling, time-domain convolutional, fully-connected and softmax layers respectively.

### 3.2.1 Feature Vector

In this study the feature vector was designed to represent an entire video. As such, the feature vector was a 3D vector with frames along one dimension, features along the second dimension, and feature components: x, y, z, and confidence along the last dimension. As a result, the feature vector on the dataset used had dimensions of: $N_{frames}* N_{joint} * N_{joint\ attributes}$ where $N_{frames}$ is the maximum number of frames in a video in the dataset, $N_{joint}$ is the number of joints associated with any given frame, and $N_{joint\ attributes}$ is the number of attributes per joint. For the processed CAD dataset: $N_{frames}$= 1961, $N_{joint}$ = 15, and $N_{joint\ attributes}$ = 4 (x, y, z, and confidence).

### 3.2.2 Neural Network Design

TensorFlow™ is open source library for numerical computation originally developed by researchers on the Google Brain Team. In this study it was used to speed the development and testing of the Neural Network. It was selected on the basis of its collection of functions relevant to machine learning and the development of neural networks, as well as its facilities for deployment of computation on various devices [30].

#### 3.2.2.1 Layers

This study used a five layer CNN as the model. The first two layers were convolutional layers, followed by two fully connected layers and finally softmax before output. The first layer of convolution processed the feature vector with a filter of dimension = 3*10 *$N_{joint\ attributes}$ and no. of kernels = 256. This is then passed to the activation function. The result of this is max pooled over joint and time dimensions. The second layer of convolution processed the output of the first layer with a filter of dimensions = 5x5x256 and no. of kernels = 64. The second layers was also followed by max-pooling over joint and time dimensions. This was followed by two fully-connected layers to allow processing across the entire feature. In both of these layers, the input was only transformed by a weight function prior to the application of the activation function. After the activation function, dropout was performed to mitigate the risk of overfitting. Dropout is a scaling of the weight variables during training to avoid having to scale down weights for testing. This relative scaling is what helps mitigate the risk of overfitting [28]. Lastly, the input is weighted and biased and then softmax is performed. Softmax regression is also known as multinomial logistic regression and can handle multiple classes. This is passed to an optimizer which minimizes the mean of this output.

#### 3.2.2.2 Initializations

For the weight initialization for each layer of the neural network, a truncated normal distribution with a low standard deviation was used to break symmetry but otherwise not impact the model to a large degree. We decided to use a slightly positive bias initialization to help avoid dead neurons due to the use of Rectified Linear Unit (ReLU) neurons. ReLU is the most popular activation function for neural networks because they alleviate the vanishing gradient problem [19]. For this reason the choice was made to run the initial model with neurons of this type.

#### 3.2.2.3 Optimizer

Multiple optimizers were considered. Many optimizers are first order gradient-based optimization methods, and some also track an estimate of various moments of the gradient to help tune step size with the goal of faster convergence [26]. An optimizer using, Adam was selected due to its good performance and convergence properties in image related neural networks when compared with AdaGrad and other techniques [24]. The formula for this optimizer is provided below in Equation 2.

$$\theta_t = \theta_{t-1} - \frac{\alpha}{\sqrt{\hat{v}_t} + \epsilon}\hat{m}_t \quad (2)$$

where,
$\theta_t$ = weight at iteration t
$\theta_{t-1}$ = weight at iteration t-1
$\alpha$ = learning rate
$\hat{m}_t$, $\hat{v}_t$ = bias corrected first and second momentum
$\epsilon$ = smoothing term (to avoid division by zero)

### 3.3 Assessments

After training the model on 80% of the dataset, the performance of the trained model was tested on the remaining data. Three metrics were selected to assess the quality of the model. These where Accuracy, Fowlkes-Mallows Index, and a Confusion Matrix.

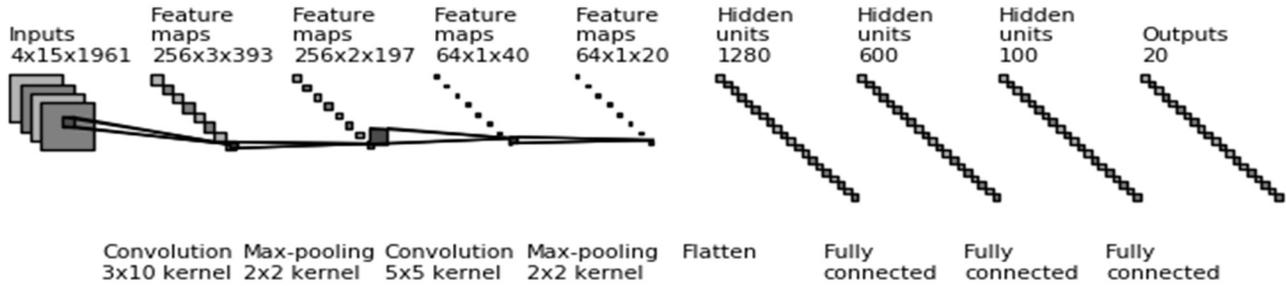

**Figure 4: Convolutional Neural Network Architecture**

Accuracy was computed by checking for the equivalence of the predicted class and the ground truth label of the class for each feature vector. If equivalence was found this was given a value of 1 and if there was no equivalence then a of 0 was given. This is equivalent to Equation 3 below. The average of these values was computed and served as the accuracy metric, as shown by Equation 4 below.

$$isEqual\left(y_{prediction}^{(i)}, y_{label}^{(i)}\right) = \begin{pmatrix} 1, & y_{prediction}^{(i)} = y_{label}^{(i)} \\ 0, & otherwise \end{pmatrix} \quad (3)$$

$$Accuracy = \frac{1}{M} \sum_{i=0}^{M-1} isEqual\left(y_{prediction}^{(i)}, y_{label}^{(i)}\right) \quad (4)$$

where,
$M$ = number of feature vectors (dataset size)
$y_{prediction}^{(i)}$ = the predicted class of a feature vector
$y_{label}^{(i)}$ = the ground truth label of a feature vector

Fowlkes-Mallows was calculated using Equation 5 below:

$$FM = \sqrt{\frac{TP}{TP+FP} \cdot \frac{TP}{TP+FN}} \quad (5)$$

where,
$TP$ = the count of true positives
$FP$ = the count of false positives
$FN$ = the count of false negatives

The confusion matrix was calculated by finding the count of the number of instances of a given prediction, label pairing and placing that value in its respective matrix coordinates, as shown in Figure 5 below.

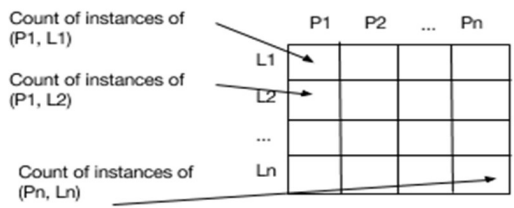

**Figure 5: Confusion Matrix Calculation.** The Pi column represents predictions in the ith class. The Li column represents labels in the i[th] class. Thus any given cell represents a specific pair of prediction and label.

## 4 Results and Discussion

The CAD-60 and CAD-120 datasets are used for testing and training our algorithms. The dimensions of the input are of the form (no.of joints * no. of frames * joint pose dimension with confidence value). The data was split into 80% for training and 20% for testing. Figure 6 compares the performance of our algorithm with different state of the art methods that uses the CAD 60 or the CAD 120 dataset. While none of the studies in the figure combined the datasets like this study did, adding complexity to the problem, they represent the majority of the corpus of work performed on the datasets. Figure 7 show the confusion matrix for the best performance obtained

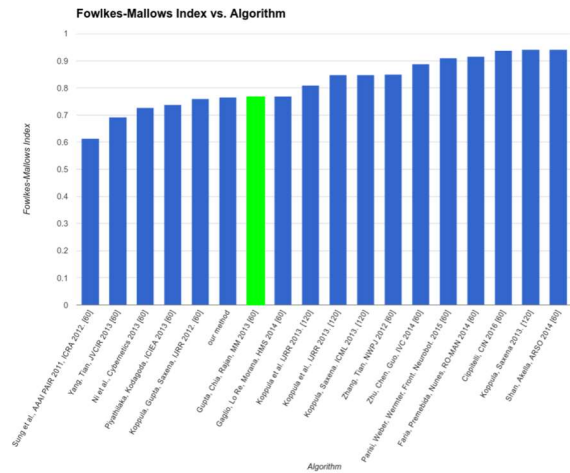

**Figure 6: Fowlkes-Mallows Index vs. Algorithm**. We can see that our model performs fairly well when compared with the state of the art results. (A bigger version attached in appendix section)

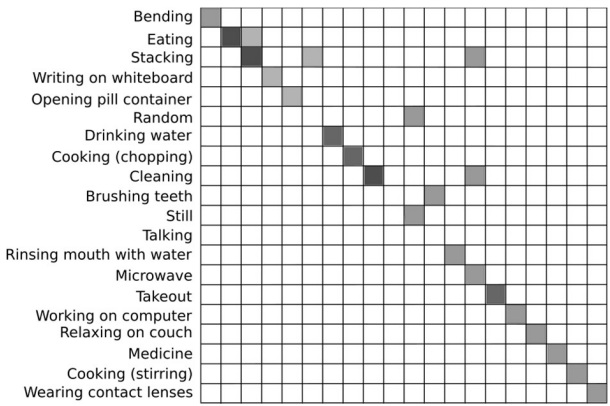

**Figure 7: Confusion matrix** of the model with accuracy 87% with Fowlkes Mallows score of 0.76

We tried using convolutional deep neural networks to model our dataset. We got a best accuracy of 70% on the testing dataset with fowlkes mallows score of 0.482.

As discussed in the data selection and preprocessing part, we planned to address the problem of less populated classes by creating new datas from the addition of a zero mean Gaussian noise to each joint in all frames for the classes that are less populated in the dataset. The standard deviation of the Gaussian noise was varied from 0.1 to 0.5 and the results are compared in the Figure 8. We could see a noticeable increase in the test accuracy as the size of datasets is increased. We got the best result of 87% with Fowlkes Mallow score of 0.76 for the Gaussian noise with standard deviation as 0.3.

We trained our model on 6 different stride lengths of the convolutional kernel (varying from 2 to 7). We saw a general trend of increasing performance until stride length 5 or 6 and then it decreased. We got the best test accuracy of 87% with Fowlkes Mallow score of 0.76 on stride length 5. The result for different stride lengths are compared in Figure 9.

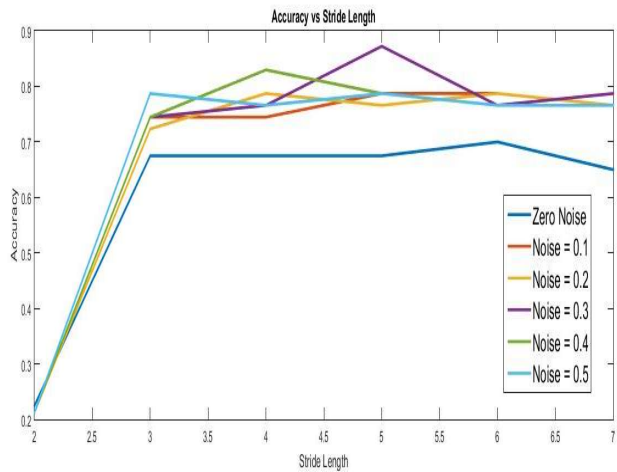

**Figure 8: Accuracy vs Stride Length**. Shows that increasing the size of data by noise addition improves the performance with the best performance observed for noise with standard deviation 0.3

We also tried using Random Forest Classifier to model our dataset. We got very poor results (test accuracy: 17.5). This might be because the number of data and the number of features were less for it to build good decision tree for our dataset.

As we could achieve a promising result with less data for our model and the model being computationally inexpensive, it encouraged us to analyze the model on different hyperparameter settings to give us a broader perspective of the model's working. We in general chose to vary the dataset size and the stride number for the convolutional kernels.

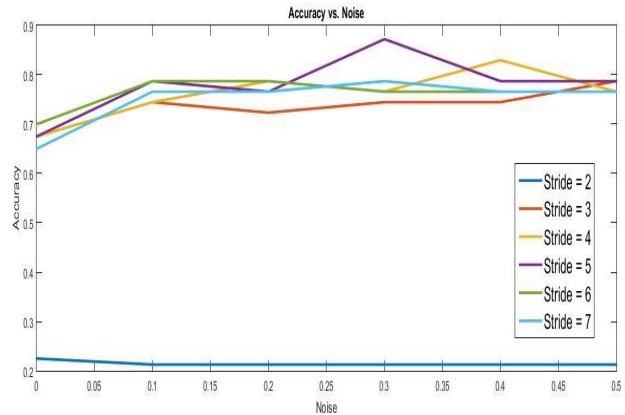

**Figure 9: Accuracy vs Noise**. Shows that the performance of our model is maximum for a stride length of 5 irrespective of the dataset used

## 5 Conclusion

Initially the dataset is classified using random forest classifier which performed only with the accuracy of 17.5%. The reason for this poor performance is the insufficient data which is required to build strong decision trees. Later we designed a convolutional neural network which is capable of classifying human actions with an accuracy of 70% and a Fowlkes Mallows Score of 0.482, when trained without Gaussian noise. When the same model is trained with a dataset containing Gaussian noise, an improvement in the performance metrics was observed. The model performed best when the dataset noise with a standard deviation of 0.3 was applied to the data. This resulted in an accuracy of 87% and a Fowlkes Mallows Score of 0.76.

The main advantage of this project is the robustness of obtaining data from different sources like kinect sensor, stereo camera, lidar, or even smart clothing as we need only the information about the joint locations. Some other advantages are, the simplicity in the model, and the accuracy the model can reach with less training data. But still the model needs some more improvement like increasing the accuracy of classifying human actions and broadening the set of actions classifiable by the network to make the model ready for implementation in real world scenarios. This can be achieved by increasing the training data. Another drawback is that not all actions can be simplified to the skeletal representation which prevents the model to learn such actions. Further investigation is needed to determine, whether the increased accuracy of the noise added data over the baseline data is due to high bias in the original training data. Similarly, additional work is needed to verify that the randomization is contained within the training only.

Further improvements and future research for this study can be focused on identifying sub-actions like talking on the phone, comparing different methods for the normalization of the number of frames in the video clips, expanding the dataset, analysing the performance when the designed model is integrated with different CNN models which can recognise the actions that cannot be represented in the skeleton form. Additional research is needed in understanding the effect of adding different types of synthetic data, like transformations of the skeleton frame by rotation and translation and different types of noise and its effects on the performance of the model.

## Acknowledgements

We would like to thank Professor Dmitry Korkin of the department of Computer Science at Worcester Polytechnic Institute for his patience, guidance, advice, and support over the term of the project, and Dr. Spencer Pruitt and the staff of the Worcester Polytechnic Institute - Academic & Research Computing (ARC) facility for permitting and facilitating our use of their Ace high-performance computing system.

## References

[1] Karpathy, A., Toderici, G., Shetty, S., Leung, T., Sukthankar, R. and Fei-Fei, L., 2014. Large-scale video classification with convolutional neural networks. In *Proceedings of the IEEE conference on Computer Vision and Pattern Recognition* (pp. 1725-1732).

[2] Vemulapalli, R., Arrate, F. and Chellappa, R., 2014. Human action recognition by representing 3d skeletons as points in a lie group. In *Proceedings of the IEEE Conference on Computer Vision and Pattern Recognition* (pp. 588-595).

[3] Koppula, H.S. and Saxena, A., 2013, June. Learning Spatio-Temporal Structure from RGB-D Videos for Human Activity Detection and Anticipation. In *ICML (3)* (pp. 792-800).

[4] Yue-Hei Ng, J., Hausknecht, M., Vijayanarasimhan, S., Vinyals, O., Monga, R. and Toderici, G., 2015. Beyond short snippets: Deep networks for video classification. In *Proceedings of the IEEE Conference on Computer Vision and Pattern Recognition* (pp. 4694-4702).

[5] A.Toshev and C. Szegedy. Deeppose: Human pose estimation via deep neural networks. In Computer Vision and Pattern Recognition (CVPR), 2014.

[6] Baccouche, M., Mamalet, F., Wolf, C., Garcia, C. and Baskurt, A., 2011, November. Sequential deep learning for human action recognition. In *International Workshop on Human Behavior Understanding* (pp. 29-39). Springer Berlin Heidelberg.

[7] Feichtenhofer, C., Pinz, A. and Zisserman, A., 2016. Convolutional Two-Stream Network Fusion for Video Action Recognition. *arXiv preprint arXiv:1604.06573*.

[8] Bilen, H., Fernando, B., Gavves, E., Vedaldi, A. and Gould, S., 2016. Dynamic image networks for action recognition. In *IEEE International Conference on Computer Vision and Pattern Recognition CVPR*.

[9] Sung, J., Ponce, C., Selman, B. and Saxena, A., 2012, May. Unstructured human activity detection from rgbd images. In *Robotics and Automation (ICRA), 2012 IEEE International Conference on* (pp. 842-849). IEEE.

[10] Gupta, A., Kembhavi, A. and Davis, L.S., 2009. Observing human-object interactions: Using spatial and functional compatibility for recognition. *IEEE Transactions on Pattern Analysis and Machine Intelligence*, *31*(10), pp.1775-1789.

[11] Du, Y., Wang, W. and Wang, L., 2015. Hierarchical recurrent neural network for skeleton based action recognition. In *Proceedings of the IEEE Conference on Computer Vision and Pattern Recognition* (pp. 1110-1118).

[12] Kwon, W.Y., Park, Y., Lee, S.H. and Suh, I.H., Human Activity Recognition Using Deep Recurrent Neural Networks and Complexity-based Motion Features.

[13] Wu, D. and Shao, L., 2014. Leveraging hierarchical parametric networks for skeletal joints based action segmentation and recognition. In *Proceedings of the IEEE Conference on Computer Vision and Pattern Recognition* (pp. 724-731).

[14] Simonyan, K. and Zisserman, A., 2014. Two-stream convolutional networks for action recognition in videos. In *Advances in Neural Information Processing Systems* (pp. 568-576).

[15] Hou, Y., Li, Z., Wang, P. and Li, W., 2016. Skeleton optical spectra based action recognition using convolutional neural networks. *IEEE Transactions on Circuits and Systems for Video Technology*.

[16] Antkowiak, M., 2006. Artificial Neural Networks vs. Support Vector machines for skin diseases recognition. *Master Degree, Department of Computing Science, Umea University, Sweden*.

[17] Banko, M. and Brill, E., 2001, March. Mitigating the paucity-of-data problem: Exploring the effect of training corpus size on classifier performance for natural language processing. In *Proceedings of the first international conference on Human language technology research* (pp. 1-5). Association for Computational Linguistics.

[18] Cippitelli, E., Gasparrini, S., Gambi, E. and Spinsante, S., 2016. A Human Activity Recognition System Using Skeleton Data from RGBD Sensors. *Computational intelligence and neuroscience*, *2016*.

[19] Clevert, D.A., Unterthiner, T. and Hochreiter, S., 2015. Fast and accurate deep network learning by exponential linear units (elus). *arXiv preprint arXiv:1511.07289*.

[20] Cornell University, 2009. *Robot Learning Lab: Cornell Activity Datasets: CAD-60 & CAD-120.* [Online] Available at: http://pr.cs.cornell.edu/humanactivities/data.php [Accessed 4 December 2016].

[21] Faria, D.R., Premebida, C. and Nunes, U., 2014, August. A probabilistic approach for human everyday activities recognition using body motion from RGB-D images. In *The 23rd IEEE International Symposium on Robot and Human Interactive Communication* (pp. 732-737). IEEE.

[22] Gaglio, S., Re, G.L. and Morana, M., 2015. Human activity recognition process using 3-D posture data. *IEEE Transactions on Human-Machine Systems*, *45*(5), pp.586-597.

[23] Koppula, H.S., Gupta, R. and Saxena, A., 2013. Learning human activities and object affordances from rgb-d videos. *The International Journal of Robotics Research*, *32*(8), pp.951-970.

[24] Kingma, D. and Ba, J., 2014. Adam: A method for stochastic optimization. arXiv preprint arXiv:1412.6980.

[25] Sapp, B. and Taskar, B., 2013. Modec: Multimodal decomposable models for human pose estimation. In *Proceedings of the IEEE Conference on Computer Vision and Pattern Recognition* (pp. 3674-3681).

[26] Sercu, T., Puhrsch, C., Kingsbury, B. and LeCun, Y., 2016, March. Very deep multilingual convolutional neural networks for LVCSR. In *2016 IEEE International Conference on Acoustics, Speech and Signal Processing (ICASSP)* (pp. 4955-4959). IEEE.

[27] Shan, J. and Akella, S., 2014, September. 3D human action segmentation and recognition using pose kinetic energy. In *2014 IEEE International Workshop on Advanced Robotics and its Social Impacts* (pp. 69-75). IEEE.

[28] Srivastava, N., Hinton, G.E., Krizhevsky, A., Sutskever, I. and Salakhutdinov, R., 2014. Dropout: a simple way to prevent neural networks from overfitting. *Journal of Machine Learning Research*, *15*(1), pp.1929-1958.

[29] TensorFlow, 2016. *TensorFlow - an Open Source Library for Machine Intelligence*. [Online] Available at: https://www.tensorflow.org/ [Accessed 4 December 2016]

[30] Zhang, C. and Tian, Y., 2012. RGB-D camera-based daily living activity recognition. *Journal of Computer Vision and Image Processing*, *2*(4), p.12.

**Appendix**

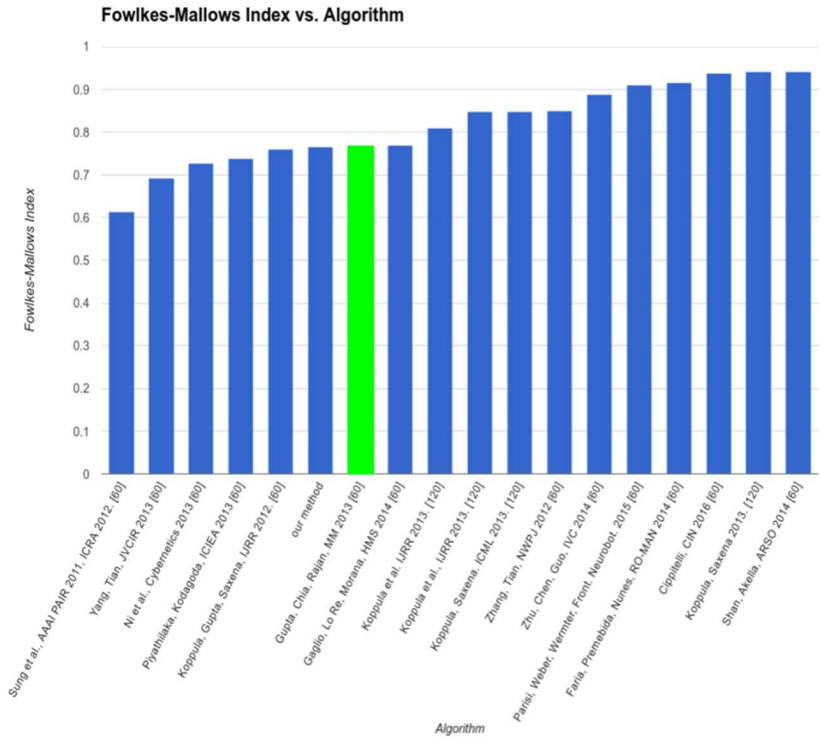

Figure 6: Fowlkes-Mallows Index vs. Algorithm